\pdfoutput=1

\documentclass[11pt]{article}

\usepackage[preprint]{acl}

\usepackage{times}
\usepackage{latexsym}

\usepackage[T1]{fontenc}

\usepackage[utf8]{inputenc}

\usepackage{microtype}

\usepackage{inconsolata}

\usepackage{graphicx}
\usepackage{xcolor}
\usepackage{subcaption}
\usepackage{pifont} 
\usepackage{amsmath}
\usepackage{amssymb}
\usepackage{cancel}
\usepackage{multirow}
\usepackage{makecell}
\usepackage{tabularx}
\usepackage{graphicx}
\usepackage{float}
\usepackage{tgtermes}
\usepackage{diagbox}
\usepackage[mathscr]{eucal}
\usepackage{amsfonts}
\usepackage[symbol]{footmisc}
\renewcommand{\thefootnote}{\fnsymbol{footnote}}
\usepackage{booktabs}
\usepackage{tablefootnote}
\usepackage{csquotes}
\usepackage{subcaption}
\usepackage{adjustbox}
\usepackage{array}
\usepackage{arydshln}
\usepackage{multicol}
\usepackage{longtable}
\usepackage{algorithm}
\usepackage[noend]{algpseudocode}
\usepackage{comment}
\usepackage{tcolorbox}
\usepackage{fancyvrb}
\definecolor{blue2}{rgb}{0.0, 0.5, 1.0}
\definecolor{mb_blue}{rgb}{0.61, 0.61, 0.98}
\definecolor{mb_red}{rgb}{1.0, 0.6, 0.6}
\definecolor{red2}{rgb}{0.82, 0.1, 0.26}
\definecolor{lightgray}{gray}{0.9}
\definecolor{DarkOrchid}{RGB}{153,50,204}
\newtcolorbox{important_blue}{
    colframe=mb_blue!50,%
    colback=mb_blue!50,%
    left=0.6pt, right=0.6pt,%
    top=1.6pt, bottom=1.6pt,%
    boxsep=0pt,%
    hbox,
    before=\vspace{0em},
    after=\vspace{0em}
}
\newtcolorbox{important_red}{
    colframe=mb_red!50,%
    colback=mb_red!50,%
    left=0.6pt, right=0.6pt,%
    top=1.6pt, bottom=1.6pt,%
    boxsep=0pt,%
    hbox,
    before=\vspace{0em},
    after=\vspace{0em}
}
\newtcolorbox{important_y}{
    colframe=y!80,%
    colback=y!80,%
    left=1pt, right=1pt,%
    top=0.5pt, bottom=0.5pt,%
    boxsep=0pt,%
    hbox,
    before=\vspace{0em},
    after=\vspace{0em}
}

\title{\raisebox{-0.1\height}{\includegraphics[height=1.6ex]{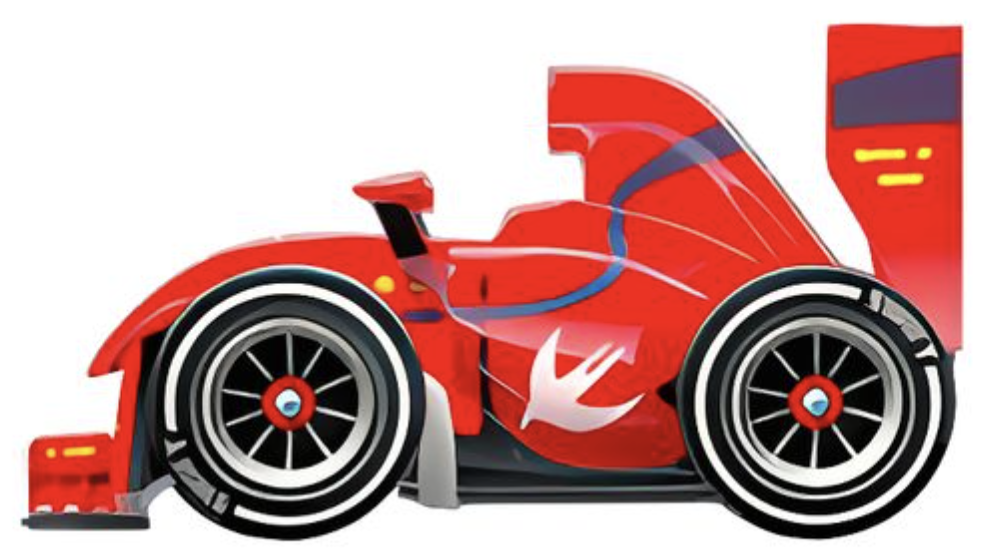}} Drift: Decoding-time Personalized Alignments with \\ \textit{Implicit} User Preferences}

\author{
Minbeom Kim$^{1\dagger}$ \hspace{1.5cm} Kang-il Lee$^{1}$ \hspace{1.5cm} Seongho Joo$^{1}$ \\ \textbf{Hwaran Lee$^{2, 3}$}  \hspace{1.5cm} \textbf{Thibaut Thonet$^{4}$}  \hspace{1.5cm} \textbf{Kyomin Jung$^{1\dagger}$} \\
    $^{1}$Seoul National University $   $\quad
    $^{2}$Sogang University $   $\quad
    $^{3}$NAVER AI Lab $  $\quad
    $^{4}$NAVER Labs Europe $  $\\
    \texttt{\{minbeomkim, kjung\}@snu.ac.kr}
}

\begin{document}
\maketitle
\begin{abstract}

Personalized alignments for individual users have been a long-standing goal in large language models (LLMs). 
We introduce \textbf{Drift}, a novel framework that personalizes LLMs at decoding time with \textit{implicit} user preferences. Traditional Reinforcement Learning from Human Feedback (RLHF) requires thousands of annotated examples and expensive gradient updates. In contrast, Drift personalizes LLMs in a \textit{training-free} manner, using \textit{only a few dozen examples} to steer a frozen model through efficient preference modeling. Our approach models user preferences as a composition of predefined, interpretable attributes and aligns them at decoding time to enable personalized generation. Experiments on both a synthetic persona dataset (\textit{Perspective}) and a real human-annotated dataset (\textit{PRISM}) demonstrate that Drift significantly outperforms RLHF baselines while using only 50–100 examples. Our results and analysis show that Drift is both computationally efficient and interpretable.


\end{abstract}

\renewcommand*{\thefootnote}{\arabic{footnote}}
\setcounter{footnote}{0}

\section{Introduction}

Large language models (LLMs) have rapidly become integral to a wide range of applications, driven by advances in \textit{Reinforcement Learning from Human Feedback} (RLHF)~\citep{ziegler2020finetune, rafailov2024direct}. Traditionally, RLHF aligns LLMs with \textbf{general preferences} by leveraging large-scale annotations from diverse users. Building on these successes, an important question naturally arises: \textit{Can we align LLMs with individual users’ \textbf{personal preferences?}}

LLM personalization, however, presents several challenges. First, collecting extensive, user-specific annotations is prohibitively expensive and impractical. Second, training and maintaining separate LLMs per user is computationally infeasible, which motivates the need for a \textit{training-free} approach. Third, while user-specific system prompts have been proposed as an alternative~\citep{hwang2023aligninglanguagemodelsuser, multifacet}, most users struggle to \textit{explicitly} articulate their complex preferences~\citep{nisbett1977telling, pronin2001you}. This often leads to misalignment between stated and actual \textit{implicit} preferences.

\begin{figure}[t!]
\centering
\includegraphics[trim=7 8 2 2, clip, width=0.78\columnwidth]{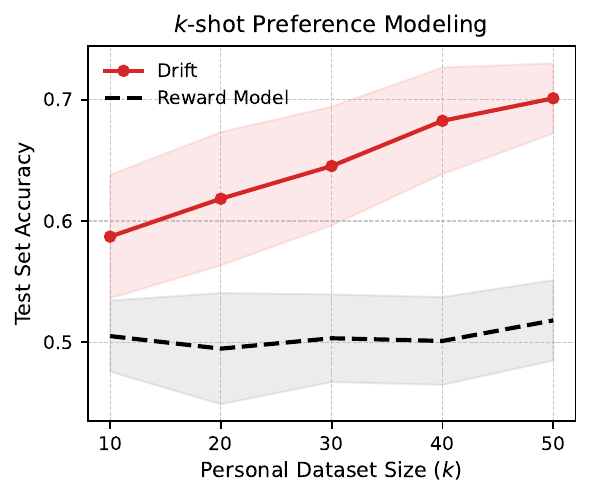}
\caption{Few-shot preference modeling for \texttt{user1008} in the PRISM dataset. 
Traditional reward models fail to generalize with scarce data, whereas Drift achieves strong prediction with only 50 examples.}
\label{fig:Intro}
\vspace{-5mm}
\end{figure}

To address these challenges, we propose \textbf{Drift}, an algorithm for few-shot personalization of LLMs that requires no gradient updates. Our key contributions are as follows:

\noindent
\textit{Drift Approximation:} We first decompose complex personal preferences into a composition of simpler and predefined attributes (e.g., ``emotional,'' ``concise,'' ``technical''). In this process, we theoretically demonstrate how to convert the RLHF objective~\citep{rafailov2024direct} into a Drift optimization problem, enabling efficient preference modeling with minimal data. 

\noindent \textit{Zero-shot Rewarding:} We leverage a \textit{differential prompting} approach to reward various attributes in a zero-shot manner. Specifically, we modify a base system prompt with attribute cues and compute the difference in log likelihood between the modified and base prompts. This differential signal acts as a surrogate reward, eliminating the need for specialized datasets or attribute-specific training.

\noindent
\textit{Drift Decoding:} Finally, using the composite preferences obtained via Drift Approximation, we derive a principled method to steer the decoding process of a frozen LLM. We prove that by integrating weighted, attribute-specific rewards into the logit space, we can achieve personalized generation without any model updates or gradient computations. 

We evaluate Drift on two fronts: (i) efficient few-shot preference modeling and (ii) personalized generation. As shown in Figure~\ref{fig:Intro}, unlike traditional reward models, Drift reaches a test-set accuracy of 70\% with only 50 examples and even outperforms a reward model trained on 500 examples. By aligning this effective preference model at decoding-time, Drift consistently produces outputs that better reflect individual users’ implicit preferences.
Extensive analysis and discussion further validate the robustness and practical benefits of Drift.


\paragraph{Contributions.}
1) We propose the first \textit{few-shot implicit personalization} and \textit{training-free} algorithm for LLMs that decomposes implicit preferences into diverse, interpretable attributes and aligns their composition at decoding-time. 2) We introduce the differential prompting technique to capture numerous aspects of personal preferences, demonstrating strong empirical results without the need for dedicated dataset construction. 3) We empirically show that Drift achieves robust few-shot preference modeling and personalized generation on both synthetic and real-world datasets, providing significant practical benefits with extensive analysis.
\section{Preliminaries}

Before describing Drift in detail, we review the standard RLHF pipeline and recent decoding-time alignment methods that motivate our approach.

\subsection{RLHF}

RLHF aims to align a base model $\pi_{\text{base}}$ with human preferences by transforming human feedback into a reward function. The typical RLHF pipeline comprises three main steps: (1) Preference data collection, (2) Preference modeling, and (3) RL training.

\paragraph{Preference data collection.} 
Given a prompt $x$, $\pi_{\text{base}}$ generates responses $(y_1, y_2, \ldots, y_t) \sim \pi_{\text{base}}(\cdot \mid x)$. Human annotators then evaluate these responses by expressing pairwise preferences, denoted as $y_w \succ y_l \mid x$, where $y_w$ is the preferred response and $y_l$ is the less preferred one. These annotated pairs form the dataset $\mathcal{D}$.

\paragraph{Preference Modeling.} The preference model (also referred to as the reward model) $r(x, y)$ is trained to capture human preferences. This is typically achieved using a Bradley-Terry loss function~\citep{bradley1952rank}:
\begin{equation}
    \max_{r} \mathbb{E}_{(x, y_w, y_l) \sim \mathcal{D}} \left[ \log \sigma\big(r(x, y_w) - r(x, y_l)\big) \right],\nonumber
\end{equation}
where $\sigma$ represents the logistic function. Through this training process, $r(x, y)$ learns to quantify the human preferences encoded in $\mathcal{D}$.

\paragraph{KL-Regularized RL.} To align the base model $\pi_{\text{base}}$ with human preferences, the objective is to maximize the reward while minimizing the KL divergence $D_{\text{KL}}$ from the base model~\citep{schulman2017proximal}, as follows:
\begin{equation}
    \max_{\pi_{\theta}} \ \mathbb{E}_{y \sim \pi_{\theta}(y \mid x)} \left[ r(x, y)  - \beta D_{\text{KL}}\big(\pi_{\theta} \| \pi_{\text{base}}\big)\right],\nonumber
\end{equation}
where $\beta$ controls the deviation ($\beta>0$).

\begin{figure*}[th]
    \centering
    \includegraphics[trim=85 62 25 66, clip, width=0.95\textwidth]{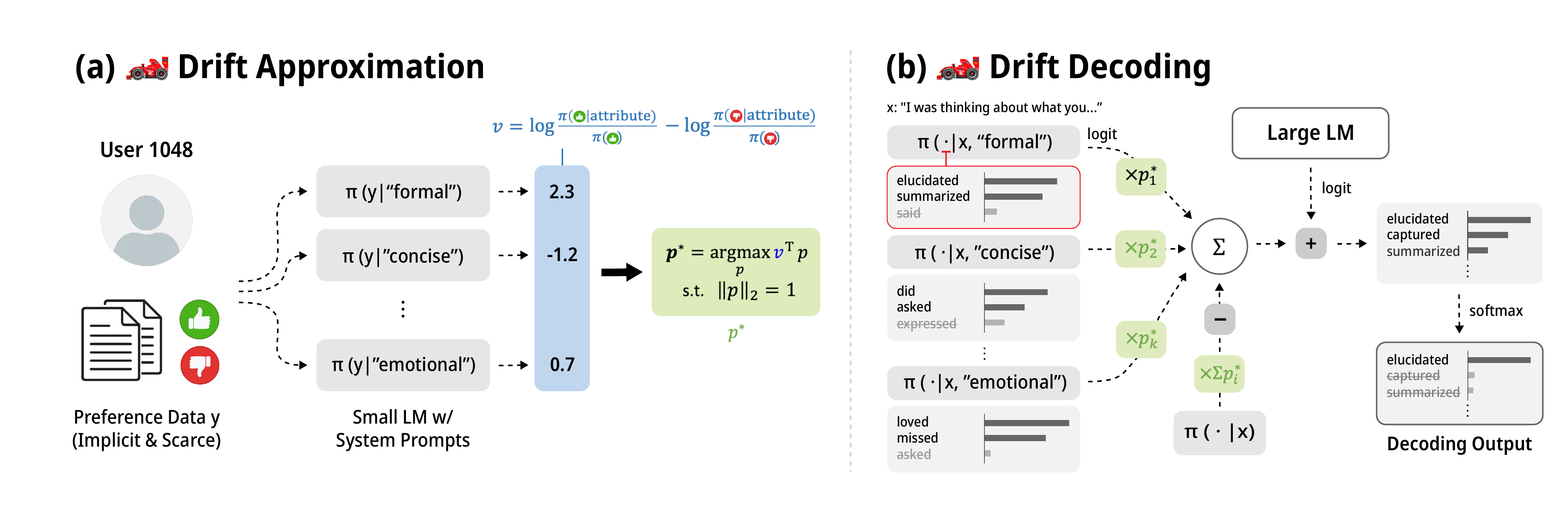}
    \caption{Overview of the total Drift Algorithms. (a) Drift Approximation: Decomposes a user’s implicit preferences into a weighted combination of various attributes. (b) Drift Decoding: Integrates this attribute composition into the decoding process without retraining the LLM.
    }
    \label{fig:main}
\vspace{-5mm}
\end{figure*}

\subsection{Decoding from RLHF Objective}

\paragraph{RL Closed-Form Solution.} The KL-regularized RL problem has a closed-form solution~\citep{korbak2022rl}:

{
\small
\begin{equation}
    \pi^*(y \mid x) = \frac{1}{Z(x)} \pi_{\text{base}}(y \mid x) \exp\left(\frac{1}{\beta} r(x, y)\right),
    \label{eq:ideal_distributions}
\end{equation}
}
where, $Z(x) = \sum_y \pi_{\text{base}}(y \mid x) \exp\left(\frac{1}{\beta} r(x, y)\right)$ is the partition function (Proof in Appendix~\ref{proof-rl}). While this solution implies the possibility of training-free alignment of $\pi_{\text{base}}$ using only $r$, in most cases, $Z(x)$ is often intractable~\citep{lin2022uncomputability}.

\paragraph{Decoding-time alignments.}
Recently, \citet{liu2024tuning, xu2024genarm, liu2024decoding} have addressed this challenge through collaborative decoding between the LLM and a smaller language model (sLM). By training a sLM $\pi_{r}$ on $\mathcal{D}$ to create an aligned model $\pi^*_{r}$, the relationship
\begin{equation}
r(x, y) = \beta\log\frac{\pi^*_r(y \mid x)}{\pi_{r}(y \mid x)} + \beta\log Z_r
\label{eq:generative_reward}
\end{equation}
is established. Although $Z_r$ remains intractable, calculations performed at the logit level yield:
\begin{equation}
\begin{split}
\pi^*(\cdot \mid x) =\; &\text{softmax}\Big( h_{\pi}(\cdot \mid x) \\
&\quad + \beta^{-1} \Bigl( h_{\pi^*_r}(\cdot \mid x) - h_{\pi_{r}}(\cdot \mid x) \Bigr) \Big).\nonumber
\end{split}
\label{eq:basic_decoding}
\end{equation}
Bypassing the computation of $Z_r$, this formulation allows practical decoding-time alignment without expensive fine-tuning.

\paragraph{Challenges for personalized alignments.}

Despite these advances, training a robust reward model typically requires large amounts of data—an impractical requirement for individual user personalization. Drift is designed to overcome this limitation via few-shot preference modeling with a weighted linear combination of various attribute-specific reward signals, which can be directly applied to decoding-time alignment.

\section{{\includegraphics[height=1.6ex]{figs/racer.png}} Drift Algorithms}

Drift overcomes data scarcity and computational inefficiency by decomposing a user’s complex personal preferences as a linear combination of simpler attributes. As Figure~\ref{fig:main}, we describe two key components: \textit{Drift Approximation}, which efficiently estimates attribute weights from a few dozen examples, and \textit{Drift Decoding}, which integrates these weights into the LLM’s decoding process.

\subsection{Drift Approximation}
\paragraph{Problem Setup.} 
Assume we have a personalized preference dataset $\mathcal{D}$, a frozen LLM $\pi_\text{LLM}$, and a set of $k$ attribute-specific small LMs $\{\pi_i^*\}_{i=1}^k$ (with corresponding base model $\pi$). We model the overall personalized reward as
\begin{equation}
    R_{\mathcal{D}}(y \mid x) = \sum_{i=1}^k p_i \, r_i(y \mid x),
\end{equation}
where $p_i$ indicates the importance of the $i$th attribute. Under the KL-regularized framework in Eq.~\ref{eq:ideal_distributions}, the target distribution $\tilde{\pi}$ becomes:

{\small
\begin{equation}
\begin{split}
    \tilde{\pi}(y \mid x) &\propto \pi_\text{LLM}(y \mid x) \exp\!\Bigl(\beta^{-1}R_{\mathcal{D}}(y \mid x)\Bigr)\\[1mm]
    &= \pi_\text{LLM}(y \mid x) \prod_{i=1}^k \exp\!\left(\frac{p_i}{\beta} \, r_i(y \mid x)\right).
    \label{eq:combined_distribution}
\end{split}
\end{equation}}
\noindent
Each reward is expressed in a generative form:
\begin{equation}
    r_i(y \mid x) = \log\!\frac{\pi_i^*(y \mid x)}{\pi(y \mid x)} + \log Z_i(x),
    \label{eq:attribute_reward}
\end{equation}
with the partition term $Z_i(x)$ canceling out in pairwise comparisons.

\paragraph{From Bradley-Terry to Drift.} 
To estimate the attributes weights $\mathbf{p} = [p_1, \dots, p_k]$, we initiate the Bradley-Terry formulation as \citet{rafailov2024direct}. For a given pair $(y_w, y_l)$ (where $y_w$ is preferred over $y_l$), we have:
\begin{align}
    &\max_{\theta} \ p(y_w > y_l \mid x) = \nonumber \\
    &\frac{1}{1 + \exp\left(\beta\left(\log\frac{\pi_\text{LLM}^{\theta}(y_l \mid x)}{\pi_\text{LLM}^{\text{ref}}(y_l \mid x)} - \log\frac{\pi_\text{LLM}^{\theta}(y_w \mid x)}{\pi_\text{LLM}^{\text{ref}}(y_w \mid x)}\right)\right)}\nonumber
\end{align}
as in DPO~\citep{rafailov2024direct}. 
Substituting Eqs.~\ref{eq:combined_distribution} and \ref{eq:attribute_reward} simplifies this optimization to:
{\tiny
\begin{align}
\max_{\mathbf{p}}\; &\frac{1}{1 + \exp\Bigl(\beta \Bigl(\sum\limits_{i=1}^k p_i \log\frac{\pi^*_i(y_l \mid x)}{\pi(y_l \mid x)} \nonumber - \sum_{i=1}^k p_i \log\frac{\pi^*_i(y_w \mid x)}{\pi(y_w \mid x)}\Bigr)\Bigr)}.
\end{align}}
By monotonicity of $x \mapsto \frac{1}{1 + \exp(-\beta x)}$, reducing the problem to a simpler optimization task:
\begin{equation}
    \max_{\mathbf{p}} \ \sum_{i=1}^k p_i \left(\log\frac{\pi_i^*(y_w \mid x)}{\pi(y_w \mid x)} - \log\frac{\pi_i^*(y_l \mid x)}{\pi(y_l \mid x)}\right).\nonumber
\end{equation}
To avoid an unbounded solution, we constrain $\mathbf{p}$ to lie on the unit $\ell_2$ sphere:
\begin{equation}
    \max_{\mathbf{p}} \ \left(\mathbf{W} - \mathbf{L}\right)^T \mathbf{p}, \quad \text{subject to } \|\mathbf{p}\|_2 = 1,
\end{equation}
where $\mathbf{W}$ and $\mathbf{L}$ aggregate the log-ratio differences for the preferred $y_w$ and less preferred $y_l$ outputs over $\mathcal{D}$, respectively. 
Notably, this approximation is completely gradient-free and thus highly efficient compared to traditional preference modeling.

\paragraph{Zero-Shot Rewarding via Differential Prompts.}  
Drift Approximation computes $r_i$ for each instance $y$ as $\log\frac{\pi_i^*(y \mid x)}{\pi(y \mid x)}$. However, training an attribute-specific model $\pi^*_i$ for every possible attribute is infeasible. Instead, we reward each attribute in a zero-shot manner using differential prompts. 

Starting from a base prompt $s_0$ (e.g., \textit{"You are an AI assistant."}), we compute the log-probability $\log \pi(y | x, s_0)$. For each attribute (e.g., \textit{\textcolor{blue2}{emotion}}), we modify the base prompt by adding a corresponding cue (e.g., \textit{"You are an \textcolor{blue2}{emotional} AI assistant."}) to obtain $s_i$ and compute $\log \pi_i^*(y | x)=\log \pi(y | x, s_i)$. Their difference $\log\frac{\pi(y \mid x, s_i)}{\pi(y \mid x, s_0)}$ captures the differential impact of the attribute cue, serving as a surrogate reward signal that measures how well the response $y$ aligns with the attribute. 
This approach is: 1) \textbf{Training-free:} No additional fine-tuning is needed, 2) \textbf{Flexible:} New attributes can be integrated on the fly, 3) \textbf{Memory efficient:} It avoids the need to maintain multiple LLMs.

Algorithm~\ref{alg:drift-approximation} summarizes the Drift Approximation procedure.

\algrenewcommand\algorithmicrequire{\textbf{Input:}}
\algrenewcommand\algorithmicensure{\textbf{Output:}}
\begin{algorithm}[t]
\caption{Drift Approximation}
\label{alg:drift-approximation}
\begin{algorithmic}[1]
\Require Dataset $\mathcal{D} = \{(y^j_w, y^j_l, x^j)\}_{j=1}^n$, sLM $\pi$, base prompt $s_0$, attribute prompts $\{s_i\}_{i=1}^k$
\Ensure Attribute weights $\mathbf{p} = \{p_1, p_2, \dots, p_k\}$
\For{$j = 1$ to $n$} \Comment{Over each data point}
    \For{$i = 1$ to $k$} \Comment{For each attribute}
        \State $\mathbf{W}_{j,i} \gets \log \frac{\pi(y^j_w\mid x^j, s_i)}{\pi(y^j_w\mid x^j, s_0)}$
        \State $\mathbf{L}_{j,i} \gets \log \frac{\pi(y^j_l\mid x^j, s_i)}{\pi(y^j_l\mid x^j, s_0)}$
    \EndFor
\EndFor
\State $\mathbf{p} \gets \arg\max_{\mathbf{p}:\|\mathbf{p}\|_2=1} \;  (\mathbf{W} - \mathbf{L})^T\mathbf{p}$
\State \Return $\mathbf{p}$
\end{algorithmic}
\end{algorithm}

\subsection{Drift Decoding}
\label{sec:drift-decoding}
Once the attribute weights $\mathbf{p}$ are obtained, Drift enables personalized generation by sampling directly from a composite distribution that adjusts the frozen LLM’s logits.
\paragraph{Composite Distribution.}  
Let $\pi_{\text{LLM}}$ denote the frozen LLM and $\{\pi_i\}_{i=1}^k$ the distributions obtained by prompting with $s_i$. Denote their respective logits by $h^{\text{LLM}}$, $h^i$, and let $h^{\text{base}}$ correspond to the base prompt $s_0$. The composite distribution $\tilde{\pi}$ of next token candidates $w$ is defined as:
\begin{equation}
    \tilde{\pi}(w) \propto \pi_{\text{LLM}}(w) \prod_{i=1}^k \left(\frac{\pi_i(w)}{\pi_{\text{base}}(w)}\right)^{\frac{p_i}{\beta}},
\end{equation}
where $\beta$ is the KL regularization hyperparameter that controls the strength of personalization.
Converting probabilities to logits (recall $\pi(w)=\text{softmax}(h[w])$ for all $w$), we obtain:
\begin{equation}
\begin{split}
    \log \tilde{\pi}(w) &= h^{\text{LLM}}[w] + \\ &\sum_{i=1}^k \frac{p_i}{\beta} \big(h^i[w] - h^{\text{base}}[w]\big) + C,
    \end{split}
\end{equation}
where $C$ is a constant independent of $w$ and will be ignored after $\text{softmax}$. Thus, sampling from $\tilde{\pi}$ amounts to:
{\small
\begin{equation}
    \tilde{\pi}(w) = \text{softmax}\Big( h^{\text{LLM}} + \sum_{i=1}^k \frac{p_i}{\beta} (h^i - h^{\text{base}}) \Big)[w].
\end{equation}}
Thus, sampling from $\tilde{\pi}$ amounts to adjusting the LLM's logits using the weighted attribute differences.
For a more detailed derivation, see Appendix~\ref{appendix:drift-decoding-proof}.

Algorithm~\ref{alg:drift-decoding} describes the complete autoregressive decoding procedure.

\begin{algorithm}[t]
\caption{Drift Decoding}
\label{alg:drift-decoding}
\begin{algorithmic}[1]
\Require Input context $x$, LLM $\pi_{\text{LLM}}$, sLM  $\pi$, base prompt $s_0$, attribute-specific prompts $\{s_i\}_{i=1}^k$, personal weights $\{p_i\}_{i=1}^k$ and strength $\beta$
\Ensure Generated sequence $y$
\State $y \gets \emptyset$
\While{not end of sequence}
    \State Compute $h^{\text{LLM}}_t \gets \pi_{\text{LLM}}(\cdot \mid x,y)$
    \State Compute $h^{\text{base}}_t \gets \pi(\cdot \mid x,y, s_0)$
    \For{$i = 1$ to $k$}
        \State Compute $h^i_t \gets \pi(\cdot \mid x,y, s_i)$
    \EndFor
    \State $h^{\text{drift}}_t \gets h^{\text{LLM}}_t + \frac{1}{\beta}\sum_{i=1}^k p_i (h^i_t - h^{\text{base}}_t)$
    \State Sample token $w_t \sim \text{softmax}(h^{\text{drift}}_t)$
    \State Append $w_t$ to $y$
\EndWhile
\State \Return $y$
\end{algorithmic}
\end{algorithm}

\paragraph{Practical Considerations.}
For Drift Approximation, a zero-shot rewarding mechanism can consider an unlimited number of candidate attributes with gradient-free computational cost. It is advantageous to evaluate as many attributes as possible, thereby increasing the likelihood that even a small, carefully selected subset will capture the full nuances of a user's preferences. In practice, we perform the approximation using a large pool of attributes (e.g., 41 candidates as detailed in Table~\ref{tab:system_prompts}) and then select a subset with the highest absolute weights $|p_i|$ for the final decoding process—our experiments ultimately use seven representative attributes. We will further discuss this in Section~\ref{sec:practical-1}.



\section{Experiments}

We evaluate Drift on two fronts: (i) efficient few-shot preference modeling and (ii) personalized generation. Our experiments are conducted on two datasets: \textbf{Perspective} (a synthetic persona dataset) and \textbf{PRISM} (an actual human-annotated dataset).

\begin{table}[ht]
\centering
\resizebox{\columnwidth}{!}{
\begin{tabular}{lccc}
\toprule
\multirow{2}{*}{\textbf{Dataset}} & \textbf{Explicit} & \textbf{Implicit persona} & \textbf{Avg. Size} \\
& \textbf{persona} & \textbf{ \textit{(Annotators)}} & \textbf{per user} \\
\midrule
PersonalLLM~\citep{personallm} & \ding{55} & Open-sourced RMs & 9,402 \\
PersonalSum~\citep{zhang2024personalsum} & \ding{51} & Human & 2.7 \\
PRISM~\citep{kirk2024prism} & \ding{51} & Human & 19.5 \\
Multifacet~\citep{multifacet} & \ding{51} & GPT-4 with persona & 3 \\
Perspective (Ours) & \ding{51} & GPT-4 with persona & 7,645 \\
\bottomrule
\end{tabular}}
\caption{Comparison of personal preference datasets. Perspective offers both scale and explicit persona information, enabling comprehensive evaluation.}
\label{tab:dataset-comparison}
\vspace{-5mm}
\end{table}

\subsection{Datasets and Evaluations.}
While actual user preferences are invaluable, collecting large-scale human preference data presents significant challenges. As shown in Table~\ref{tab:dataset-comparison}, existing human-annotated datasets typically contain only a few examples per user, making it difficult to train reliable reward models for evaluation. Moreover, when evaluating generation tasks, it's practically impossible to find annotators to evaluate newly generated outputs again. To address these issues, we first experiment with preference modeling and personalized generation on synthetic personas and then validate our findings using real-world data.

\paragraph{Perspective.}~\label{sec:perspective-advantages}
We introduce \textit{Perspective}, a large-scale dataset that employs synthetic personas with diverse viewpoints for reliable evaluation. Personas are selected from the Multifacet dataset~\citep{multifacet}, and we curate corresponding QA pairs that GPT-4 annotates according to each persona (details in Appendix~\ref{sec:perspective-details}). With an average of 7,645 examples per persona, Perspective offers two key advantages:
\begin{enumerate} 
\item The abundance of examples per persona allows us to train gold reward models that serve as dependable evaluation metrics for both approximation and generation tasks. 
\item The explicit persona information and consistent annotation process facilitate controlled and repeatable evaluations of generation tasks. 
\end{enumerate}

\paragraph{PRISM.}
In contrast, the PRISM dataset comprises human-annotated preferences, averaging 19.5 examples per user. We use PRISM to validate Drift’s performance in real-world scenarios, particularly under conditions of limited user data. For our experiments, we selected six users with more than 50 annotated pairs each and conducted few-shot personalization experiments to assess Drift's practical effectiveness.

\subsection{Few-shot Preference Modeling}

\begin{figure*}[th]
    \centering
        \begin{subfigure}{0.24\textwidth}
            \centering
            \includegraphics[width=\textwidth]{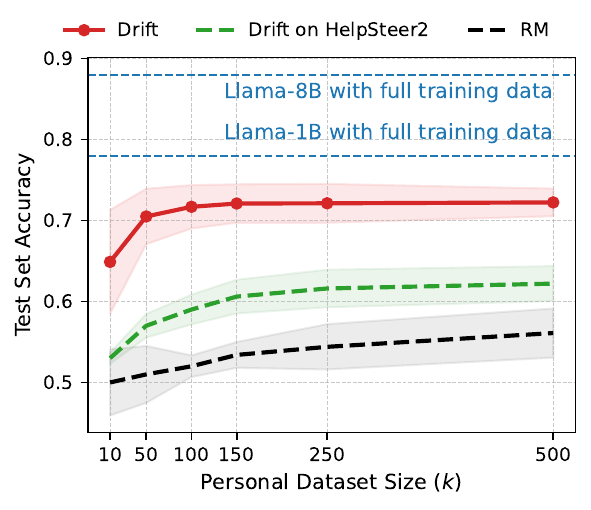}
        \end{subfigure}
        \begin{subfigure}{0.24\textwidth}
            \centering
            \includegraphics[width=\textwidth]{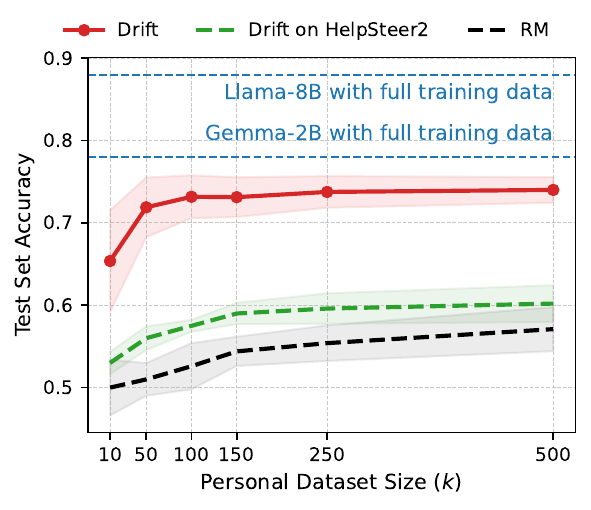}
        \end{subfigure}
        \begin{subfigure}{0.24\textwidth}
            \centering
            \includegraphics[width=\textwidth]{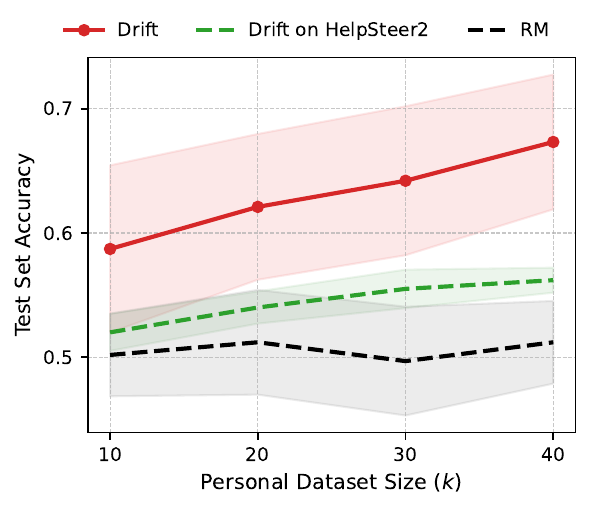}
        \end{subfigure}
        \begin{subfigure}{0.24\textwidth}
            \centering
            \includegraphics[width=\textwidth]{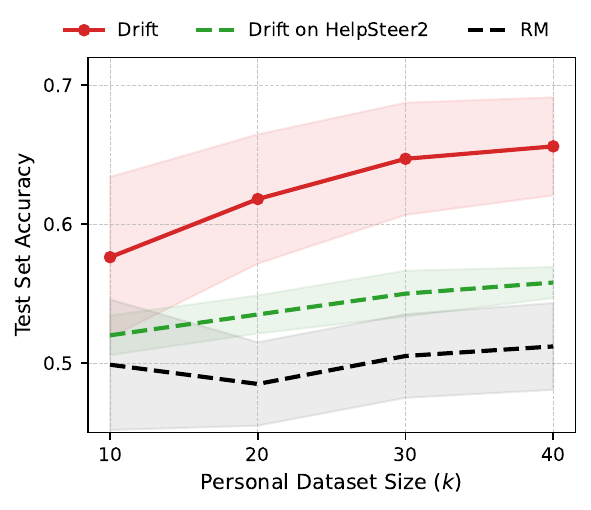}
        \end{subfigure}
    \caption{
    Average $k$-shot preference modeling results across personas in the Perspective and PRISM datasets. The two figures on the left show the results for Perspective using Llama 1B and Gemma 2B; the two on the right for PRISM using Llama 1B and Gemma 2B.
    }
    \label{fig:rm-results}
\vspace{-5mm}
\end{figure*}

\paragraph{Experimental setting.}
We evaluate the efficiency of the Drift approximation on both datasets. For Perspective, we vary the training set size from 10 to 500 examples; for PRISM, from 10 to 40 examples. Drift is compared against traditional reward models (RMs) implemented using Llama-1B~\citep{llama} and Gemma-2B~\citep{gemma}. A Llama-8B model trained on the full dataset (Gold RM) serves as an upper bound. Additionally, to assess the benefits of \textit{differential prompting}, we conduct experiments using the Drift approximation on Helpsteer2~\citep{helpsteer}, which provides well-defined attributes through specifically constructed datasets—"helpfulness," "correctness," "coherence," "complexity," and "verbosity."

\paragraph{Results.}
Figure~\ref{fig:rm-results} demonstrates that the Gold RM achieves nearly 80\% accuracy on the test set (with 8B exceeding 85\%) when trained on extensive data, while the performance of standard RMs drops below 60\% when fewer than 500 examples are available. In contrast, Drift achieves superior performance using only 50 samples, outperforming an RM trained on 500 examples with lower variance. Performance improves sharply with 100 examples and plateaus thereafter, although predictive stability continues to increase. Moreover, in PRISM involving actual users, RMs perform nearly at random, whereas Drift maintains robust approximation capabilities with just 40 samples. 
Furthermore, while Helpsteer2 offers precise reward signals for its five well-defined attributes, its limited scope fails to represent the richness of individual user preferences. Consequently, Helpsteer2 fails to generalize as effectively as our zero-shot rewarding approach, leading to lower performance compared to Drift.
These results underscore Drift’s strong \textit{few-shot personalization} capabilities, demonstrating that decomposing implicit personal preferences into multiple attributes via our \textit{differential prompting} yields robust modeling even under data scarcity.

\begin{figure}[ht]
\centering
\includegraphics[trim=7 8 2 2, clip, width=0.7\columnwidth]{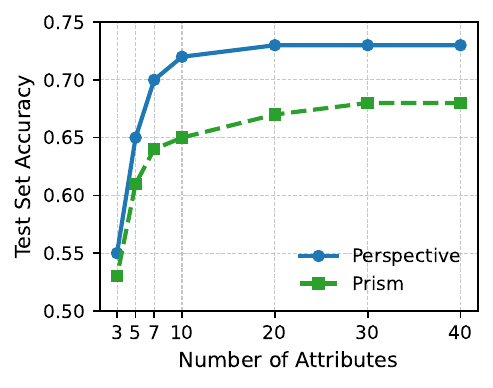}
\caption{Performance variation when reducing the number of attributes during Drift Approximation with 40 samples. The performance decline is slightly more pronounced in the PRISM dataset, suggesting that real users’ implicit preferences are more complex than those of synthetic personas.}

\label{fig:attributes_num}
\vspace{-5mm}
\end{figure}

\paragraph{Attribute Reduction for Decoding.}
\label{sec:practical-1}
Although Drift initially employs a large pool of candidate attributes (e.g., 40), only a subset is used during decoding. Figure~\ref{fig:attributes_num} shows that reducing the attribute count from 40 to 10 incurs only a modest drop in performance. Even with five attributes, the performance is significantly better than that of HelpSteer2. This suggests that a few core attributes selected by zero-shot rewarding suffice to capture personal preferences effectively. By evaluating a wide variety of attributes during the cost-efficient approximation stage, Drift identifies the most informative attributes for efficient decoding without compromising overall performance.

\paragraph{Interpretability.}
During the Drift Approximation process, we compute the average reward $W$ assigned to win responses and $L$ assigned to lose responses for each attribute. The difference $W-L$ for each attribute can be interpreted as \textbf{Unit implicit preference}, a measure of how strongly an individual implicitly prioritizes each attribute. For example, in PRISM, \texttt{user1280} introduced themselves as someone who \textit{uses an LLM exclusively for language learning}. The activated attributes reveal that ``concise" is highly prioritized with a value of 1.46, while attributes such as ``old-fashioned," ``exclamatory," and ``proverb"—which could potentially hinder language learning—are least preferred, scoring -1.19, -1.10, and -1.09, respectively. Thus, Drift not only delivers effective preference modeling with a few dozen examples but also provides valuable interpretability at the user level (additional analysis is provided in Appendix~\ref{app:attributes-activation}).

\subsection{Personalized Generation}

Next, we validate how Drift's robust preference model can be effectively integrated into personalized generation.

\paragraph{Experimental setting.}
We evaluate personalized generation primarily on the Perspective dataset, which offers reliable persona-specific evaluation metrics via a Gold RM and a GPT-based judge (see Section~\ref{sec:perspective-advantages}). Our evaluation compares the win rate of each baseline output against pure LLM outputs using these metrics. Under a few-shot setting with 100 training examples, Drift decoding is compared against several baselines:
\begin{enumerate}
    \item Training-base: PPO~\citep{schulman2017proximal}, DPO~\citep{rafailov2024direct}, DPA~\footnote{DPA training uses the weight from Drift on Helpsteer2.}~\citep{DPA}
    \item Training-free (as Drift): ARGS~\citep{khanov2024args}, Best-of-$N$ sampling ($N$=10)
\end{enumerate}
We use Llama-8B (controlled by Llama-1B as the RM) and Gemma-9B (controlled by Gemma-2B) for model configurations. Due to PRISM’s limited evaluation capabilities, we complement our quantitative results with qualitative case studies involving actual users in PRISM.

\begin{table}[th]
    \centering\resizebox{\columnwidth}{!}{
    \begin{tabular}{lccccc}
        \toprule
        \multirow{2}{*}{Method} & \multicolumn{2}{c}{Llama-8B} & & \multicolumn{2}{c}{Gemma-9B} \\
        \cmidrule(lr){2-3} \cmidrule(lr){5-6}
        & Gold RM & GPT-Judge & & Gold RM & GPT-Judge \\
        \midrule
        \multicolumn{6}{l}{\textit{Training-based}} \\
        PPO & 0.48 & 0.45 & & 0.47 & 0.46 \\
        DPO & 0.53 & 0.52 & & 0.52 & 0.51 \\
        DPA & 0.55 & 0.56 & & 0.56 & 0.56 \\
        \midrule
        \multicolumn{6}{l}{\textit{Training-free}} \\
        ARGS & 0.51 & 0.50 & & 0.51 & 0.51 \\
        Best-of-N & 0.53 & 0.54 & & 0.52 & 0.53 \\
        Drift (Ours) & \textbf{0.61} & \textbf{0.63} & & \textbf{0.62} & \textbf{0.63} \\
        \bottomrule
    \end{tabular}}
    \caption{Experimental results of \textit{Personalized Generation}. \textbf{Bold} font indicates the highest score.}
    \label{tab:pg-results}
    \vspace{-5mm}
\end{table}

\paragraph{Results.}
Table~\ref{tab:pg-results} summarizes the few-shot personalized generation results with 100 training sets from the Perspective dataset. As observed in the previous section, the RM exhibits significant shortcomings with limited data. Consequently, PPO—which relies on the reward—experiences a performance drop. In contrast, DPO, which does not depend directly on the reward signal, avoids this decline; however, its performance improvement remains marginal. Although DPA leverages a robust preference model with HelpSteer2, its effect is also limited.
Among the training-free methods, both ARGS and Best-of-$N$ sampling are more robust than the PPO because they avoid unstable training with noisy RM. Nonetheless, since they rely on the RM signal for search, their performance gains are limited.
In contrast, Drift decodes by leveraging efficient and robust approximations, resulting in significant performance improvements across all models and evaluation metrics. These findings confirm that effective \textit{few-shot personalized generation} can be achieved with as few as 100 training examples using Drift in a training-free manner.

\paragraph{Case Study in PRISM.}
We compare the user-preferred and rejected responses from PRISM with the Drift outputs. In Table~\ref{tab:case-study}, \texttt{user1008} prefers responses that present diverse opposing viewpoints rather than neutral opinions on sensitive topics. Drift effectively guides the LLM in generating responses aligned with this user preference, moving away from generic neutral responses. This demonstrates that Drift achieves robust personalization even in few-shot scenarios. The full version and additional case studies are provided in Appendix~\ref{app:prism_analysis_section}.

\begin{table}[t]
\centering
\small
\begin{tabular}{p{7cm}}
\toprule
\textbf{User Query:} \textit{"I need information about the Civil War, a topic that has caused much disagreement."} \\
\midrule
\textbf{Win Response} \\
"Sure! I can provide some basic information about the Civil War, which may help clarify the 
\textbf{\textcolor{red2}{differences of opinion people may have}}... " \\
\midrule
\textbf{Lose Response} \\
"Sure, I'd be happy to help you ... As a language model, \textbf{\textcolor{blue2}{I do not have personal opinions or biases towards any topics or discussions}}, including the Civil War." \\
\midrule
\textbf{Llama-8B with Drift Response:} \\
"A sensitive and complex topic indeed. The American Civil War, fought from ... \textbf{\textcolor{red2}{What specific aspects of the Civil War would you like to know more about?}}" \\
\midrule
\textbf{Llama-8B Pure Response:} \\
"... \textbf{\textcolor{blue2}{I'll try to provide a neutral and fact-based overview of the topic.}}" \\
\bottomrule
\end{tabular}
\caption{This table contains the \texttt{user1008} question from PRISM, with the \textbf{\textcolor{red2}{selected}} and \textbf{\textcolor{blue2}{rejected}} answers and the responses from Llama-8B w/ and w/o Drift decoding.}
\label{tab:case-study}
\vspace{-3mm}
\end{table}

\paragraph{Inference cost.}

\begin{table}[htbp]
  \centering\resizebox{0.7\columnwidth}{!}{
  \begin{tabular}{ll}
    \toprule
    Method & Time complexity \\
    \midrule
    Best-of-$N$ & $T(N\cdot C + N\cdot c)$ \\
    ARGS (top-$N$ tokens) & $T(C + N\cdot c)$ \\
    Drift ($N$ attributes) & $T(C + N\cdot c)$ \\
    \bottomrule
  \end{tabular}}
  \caption{Time complexity of each training-free method. Here, $C$ represents the LLM inference cost, $c$ denotes the sLM inference cost, and $N$ is the key hyperparameter for each method.}
  \label{tab:time_complexity}
  \vspace{-5mm}
\end{table}

Table~\ref{tab:time_complexity} summarizes the time complexity of each training-free baseline. 
Best-of-$N$ involves sampling from the LLM $N$ times and evaluating each sample with the sLM, resulting in a total complexity of $T(N\cdot C + N\cdot c)$. This method is the most computationally expensive. Both ARGS and Drift steer the LLM's next-token distribution using sLM. However, while ARGS sequentially samples from LLM and evaluates top-$N$ tokens, Drift samples the next-token distributions in parallel both from the LLM and sLM and then combines them. This flexibility offers improved efficiency over ARGS under the same memory size.

\section{Discussion}

\begin{figure}[ht]
\centering
\includegraphics[trim=7 8 2 2, clip, width=0.65\columnwidth]{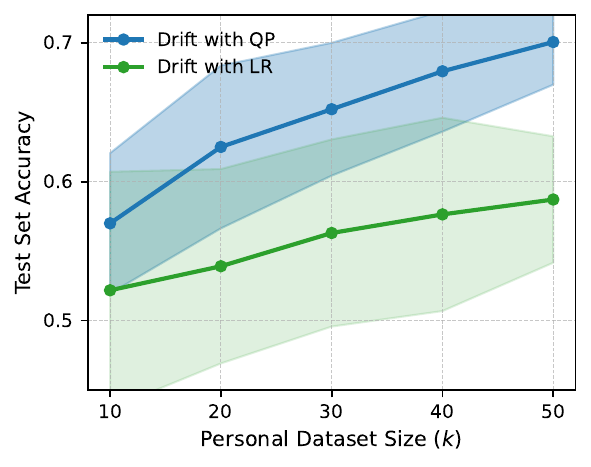}
\caption{Few-shot preference modeling results for \texttt{user1008} in the PRISM with quadratic programming (QP) and logistic regression (LQ).}
\label{fig:discussion}
\vspace{-5mm}
\end{figure}

\paragraph{Quadratic Programming vs. Logistic Regression.}  
Our formulation estimates the attribute weights $\mathbf{p}$ by transforming the Bradley-Terry loss into a quadratic program. An alternative approach based on logistic regression—which assigns absolute labels of 1 and 0 to win/lose responses—can also be used, as demonstrated by \citep{go2023compositional}. 
We compared these two formulations using Drift attributes in Table~\ref{fig:discussion}. The logistic regression approach proves highly unstable and shows lower performance when training examples are limited. We interpret this instability as follows: preference judgments are inherently relative—what constitutes a winning response in one context might be considered a losing response when compared to an even better alternative. Thus, imposing absolute labels through regression can lead to overfitting, particularly when data are scarce. Our results suggest that approaching preference problems from a relative perspective is crucial for effective preference modeling.

\paragraph{Compatible with samplers.}
\label{sec:practical-2}
Autoregressive sampling in LLMs has various decoding strategies at the token-level distribution. Drift steers distributions at the logit level—applying its computations before the softmax—making it compatible with a wide range of sampling methods tailored to different objectives~\citep{vijayakumar2016diverse, fan2018hierarchical, holtzman2019curious}. our analysis indicates that the backbone LLM exhibits an average next-token entropy of about 0.27 bits, which increases to approximately 0.63 bits after applying Drift. While this boost in entropy can substantially enhance generation diversity, it may also increase the likelihood of selecting unreliable tokens. Therefore, we recommend combining Drift with top-p or top-k sampling strategies to control an optimal balance between diversity and reliability.

\paragraph{Practical Implications.}
While traditional RLHF methods may eventually surpass Drift when user data becomes abundant, Drift offers several advantages in practical settings. 
First, conventional reward models struggle with \textit{continual learning}; retraining on an ever-expanding user dataset is impractical. In contrast, Drift can be updated instantly by simply appending new instances to the $W-L$—no retraining required. 
Second, personal preferences often \textit{change more rapidly than general preferences}. Drift’s interpretability allows real-time tracking of preference shifts, enabling dynamic adjustments for improved personalization. 
Third, when collecting additional user annotations, the variance observed in each attribute can inform an \textit{active learning} strategy~\citep{miller2020active} for efficient data collection. These benefits make Drift an attractive complement to existing RLHF pipelines in personalized applications.

\section{Related Works}
\paragraph{Explicit Personalization.}
As humans express their own preferences, recent works explored aligning LLMs with individual values through explicit cues. Multifacet~\citep{multifacet} has focused on designing diverse and detailed system prompts for LLM control. PAD~\citep{chen2024pad} and MetaAligner~\citep{yang2024metaaligner} have leveraged fine-grained RM—such as HelpSteer~\citep{helpsteer}—to construct specific policies and guide model behavior toward system prompts. Others allow users to directly specify attribute importance weights, either for training~\citep{yang2024rewards, DPA} or decoding-time alignments~\citep{dekoninck2023controlled, shi2406decoding}. 

\paragraph{Implicit Personalization.}
While they have advanced explicit preferences, implicit preferences behind users' behaviors remain understudied, as Table~\ref{tab:method_comparison}. \citet{jin2024implicit} has shown that these values arise from complex interactions between factors like experiences, education, lifestyle, and even dietary habits, leading to misalignment with explicitly stated preferences~\citep{nisbett1977telling}
To address this gap, several works proposed implicit personalization tasks - from title generation~\citep{ao-etal-2021-pens}, movie tagging~\citep{salemi2023lamp} to summarization~\citep{zhang2024personalsum}.
Notably, PRISM~\citep{kirk2024prism} made notable progress by collecting preference annotations from conversations with over a thousand users, though its effectiveness was limited by the small number of annotations per user, making traditional RLHF approaches challenging.

Our work advances this field in two key ways: First, we introduce the Perspective dataset, which enables more reliable evaluation. Second, we propose Drift, \textit{decoding-time few-shot personalization}. 
By addressing the challenges of implicit preferences, our approach represents a significant step forward in implicit personalized alignments.
\section{Conclusion}

We introduce \textbf{Drift}, a training-free framework for personalizing LLMs via decoding-time alignment with implicit user preferences. By decomposing implicit personal preferences into a weighted combination of interpretable attributes, Drift enables few-shot personalization that is both computationally efficient and interpretable. 
In the personalized AI domain—where obtaining sufficient annotations is challenging and training individual, user-specific LLMs is impractical—Drift represents a significant advance. Ultimately, Drift holds promise for democratizing personalized AI services and fostering human-aligned AI systems that are adaptable to individual needs.


\section*{Limitations}

While Drift contributes promising advances in implicit personal preferences, several limitations remain that should be addressed in future research.

\paragraph{Needs of Online Human Evaluation Benchmarks.}  
A major challenge in personal preference research is the absence of reproducible human evaluations. Even if future benchmarks collect more user-specific annotations beyond PRISM, evaluating personalized generation outputs requires \textit{re-engaging with the same users for feedback}. Although we designed the Perspective dataset to align the label construction and test set evaluation pipelines, it still relies on virtual personas. Therefore, to advance this field, there is a need for online evaluation benchmarks that can reproducibly assess personalized generation using real user feedback.

\paragraph{Limited Analysis Between Drift Attributes and Actual Users.}  
Due to practical and ethical issues, we do not have full access to the backgrounds of actual users. While the PRISM dataset provides basic information (e.g., the intended use of LLMs and brief self-introductions), our analysis (as seen in Figure~\ref{fig:activated-attributes-prism}) is limited in explaining why certain attributes are activated and how these relate to user characteristics. A more in-depth investigation into the correlation between Drift attributes and real user profiles should be studied with future benchmarks.

\paragraph{Biases in Differential Prompting.}  
Our study does not thoroughly analyze the limitations of the zero-shot rewarding mechanism used for each attribute. It is possible that differential prompting may fail to capture certain attributes accurately, and methods like those employed in Helpsteer2—where data is explicitly constructed—could offer more precise evaluations. Nevertheless, given the vast diversity of personal preferences, a zero-shot approach remains essential. As shown in Figure~\ref{fig:rm-results}, this approach yields significantly higher performance, and Figure~\ref{fig:attributes_num} demonstrates that even when the number of attributes is reduced to levels comparable to those used in Helpsteer2, performance remains robust. In essence, unreliable attributes are unlikely to be used during decoding, which mitigates this limitation. Moreover, as future research develops to enable LLM to follow system prompts more precisely, these advances will directly enhance Drift.

\paragraph{Tokenizer Dependency.}  
Drift Decoding adjusts the next-token distribution at each step, which requires that the LLM and the sLM share the same support—that is, they must use the same tokenizer. 

\paragraph{Limited Baselines.}  
Due to the scarcity of datasets for implicit personal preferences, this domain is far less mature compared to explicit preferences. As highlighted in Table~\ref{tab:method_comparison}, the limited availability of extensive baselines forced us to concentrate primarily on analyzing the unique characteristics of Drift.

\section*{Ethical Statement}

While Drift effectively integrates users’ implicit preferences into generated outputs, it also introduces several ethical risks that must be carefully managed. Notably, the Drift Approximation stage allows us to directly assess the activation levels of each attribute, which provides an opportunity to identify and preemptively block system prompts that could lead to harmful or undesirable content before they are incorporated into the decoding process. This capability underscores the importance of further research into combining Drift with diverse system prompts, ensuring that the generation of undesirable content is minimized while still delivering personalized services.

Additionally, considering that existing research~\citep{kim2024guaranteed} indicates it is impossible to obtain filtered autoregressive distributions under certain conditions, it is necessary to combine rejection sampling on final outputs using safeguards~\citep{lifetox} such as LlamaGuard~\citep{llamaguard} and ShieldGemma~\citep{shieldgemma}. This approach can further enhance the safety of the final generated content.

\bibliography{custom}

\appendix

\clearpage

\begin{table*}[t]
\centering
\resizebox{\textwidth}{!}{
\begin{tabular}{l|cccc} 
\toprule
Method & Training-free & General Policy & Smaller LM Guidance & Implicit Preference \\ 
\midrule
MORLHF~\citep{li2020deep}        & \textcolor{red}{$\times$}  & \textcolor{green}{$\checkmark$} & - & \textcolor{green}{$\checkmark$}  \\
MODPO~\citep{zhou2024beyond}   & \textcolor{red}{$\times$}      & \textcolor{green}{$\checkmark$} & -   & \textcolor{green}{$\checkmark$}  \\
Personalized soups~\citep{jang2023personalized}  & \textcolor{red}{$\times$}& \textcolor{red}{$\times$}  & \textcolor{red}{$\times$}  & \textcolor{red}{$\times$} \\
Preference Prompting~\citep{jang2023personalized} &  \textcolor{green}{$\checkmark$}  &\textcolor{green}{$\checkmark$}  & -  & \textcolor{red}{$\times$} \\
Rewarded soups~\citep{rame2024rewarded}   & \textcolor{red}{$\times$} & \textcolor{red}{$\times$}  & \textcolor{red}{$\times$}  & \textcolor{red}{$\times$}  \\
RiC~\citep{yang2024rewards}    & \textcolor{red}{$\times$}       & -  & \textcolor{red}{$\times$} & \textcolor{red}{$\times$}  \\
DPA~\citep{DPA}     & \textcolor{red}{$\times$}      & \textcolor{green}{$\checkmark$} & - & \textcolor{red}{$\times$}  \\
ARGS~\citep{khanov2024args}   & \textcolor{green}{$\checkmark$}       & \textcolor{green}{$\checkmark$}  & \textcolor{green}{$\checkmark$} & \textcolor{green}{$\checkmark$}  \\
MOD~\citep{shi2406decoding}    & \textcolor{green}{$\checkmark$}       & \textcolor{red}{$\times$}  & \textcolor{red}{$\times$}   & \textcolor{red}{$\times$}  \\
MetaAligner~\citep{yang2024metaaligner}  & \textcolor{green}{$\checkmark$}  & \textcolor{green}{$\checkmark$}  & \textcolor{green}{$\checkmark$}   & \textcolor{red}{$\times$} \\
PAD~\citep{chen2024pad}  & \textcolor{green}{$\checkmark$}  & \textcolor{green}{$\checkmark$}  & \textcolor{red}{$\times$}   & \textcolor{red}{$\times$} \\
\midrule
\includegraphics[height=1.8ex]{figs/racer.png} Drift~(Ours)   & \textcolor{green}{$\checkmark$}         & \textcolor{green}{$\checkmark$} & \textcolor{green}{$\checkmark$} & \textcolor{green}{$\checkmark$} \\
\bottomrule
\end{tabular}
}
\caption{Key characteristics of previous methods and Drift.} 
\label{tab:method_comparison}
\end{table*}

\section{Proof}

\subsection{Derivation of the RL Closed-Form Solution}
\label{proof-rl}

We want to solve the following optimization problem (for a single variable \(x\)):
\[
\max_{\theta} \Bigl[\,r(x)\;-\;\beta \,\log \tfrac{\pi_{\theta}(x)}{\pi_{base}(x)}\Bigr].
\]
Define \(\pi(x) = \pi_{\theta}(x)\). The quantity we want to maximize can be thought of as an expectation under \(\pi(x)\):
\[
\max_{\pi} \int \pi(x) \Bigl[r(x) - \beta\,\log \tfrac{\pi(x)}{\pi_{base}(x)}\Bigr] \, dx,
\]
subject to
\[
\int \pi(x)\, dx = 1 \quad\text{and}\quad \pi(x) \ge 0.
\]

Introduce a Lagrange multiplier \(\lambda\) to enforce the normalization constraint \(\int \pi(x)\,dx = 1\). The Lagrangian is
\begin{equation}
\begin{split}
    \mathcal{L}[\pi, \lambda] 
= &\int \pi(x)\,\Bigl[r(x) - \beta\,\log \tfrac{\pi(x)}{\pi_{base}(x)}\Bigr]\,dx 
\;\\- &\;\lambda\!\Bigl(\int \pi(x)\,dx - 1\Bigr). \nonumber
\end{split}
\end{equation}

We now take the functional derivative of \(\mathcal{L}\) w.r.t.\ \(\pi(x)\) and set it to zero for optimality:
\[
\frac{\delta \mathcal{L}}{\delta \pi(x)} 
= r(x) 
- \beta \Bigl[\log \tfrac{\pi(x)}{\pi_{base}(x)} + 1 \Bigr]
- \lambda 
= 0.
\]

Rearranging:
\[
r(x) - \beta \,\log \tfrac{\pi(x)}{\pi_{base}(x)} - \beta - \lambda = 0,
\]
which implies
\[
\beta \,\log \tfrac{\pi(x)}{\pi_{base}(x)} = r(x) - \beta - \lambda.
\]

Exponentiate both sides:
\[
\tfrac{\pi(x)}{\pi_{base}(x)} 
= \exp\!\Bigl(\tfrac{r(x)}{\beta}\Bigr) \;\exp\!\Bigl(- 1 - \tfrac{\lambda}{\beta}\Bigr).
\]
So
\[
\pi(x) 
= \pi_{base}(x)\,\exp\!\Bigl(\tfrac{r(x)}{\beta}\Bigr)\,\exp\!\Bigl(-1 - \tfrac{\lambda}{\beta}\Bigr).
\]
Let $C = \exp\!\Bigl(-1 - \tfrac{\lambda}{\beta}\Bigr)$.
Hence
\[
\pi(x) = C\,\pi_{base}(x)\,\exp\!\Bigl(\tfrac{r(x)}{\beta}\Bigr).
\]

We find \(C\) by imposing the constraint \(\int \pi(x)\,dx = 1\):
\[
1 
= \int \pi(x)\,dx 
= C \int \pi_{base}(x)\,\exp\!\Bigl(\tfrac{r(x)}{\beta}\Bigr)\,dx.
\]
Therefore
\[
C 
= \biggl[\int \pi_{base}(x)\,\exp\!\Bigl(\tfrac{r(x)}{\beta}\Bigr)\,dx \biggr]^{-1}.
\]

Putting it all together, the optimal distribution \(\pi^{*}(x)\) is
\[
\pi^{*}(x) 
= \frac{\pi_{base}(x)\,\exp\!\Bigl(\tfrac{r(x)}{\beta}\Bigr)}
       {\displaystyle\int \pi_{base}(x)\,\exp\!\Bigl(\tfrac{r(x)}{\beta}\Bigr)\,dx}.
\]

This shows that the optimal solution is a \emph{Boltzmann-like} (or \emph{softmax}) distribution given by weighting the reference distribution \(\pi_{base}(x)\) with the exponential of the scaled reward \(r(x)/\beta\).

\subsection{Expanded Explanation for Drift Decoding}
\label{appendix:drift-decoding-proof}
In Section~\ref{sec:drift-decoding}, we introduced the following target distribution for Drift Decoding:
\begin{equation}
    \label{eq:drift-decoding-target-dist-appx}
    \tilde{\pi}(w) \;\propto\; \pi_{\text{LLM}}(w) 
    \;\prod_{i=1}^{k} \Bigl(\tfrac{\pi_i(w)}{\pi_{\text{base}}(w)}\Bigr)^{\frac{p_i}{\beta}},
\end{equation}
where \(\pi_{\text{LLM}}(w)\) is the probability of token \(w\) under the LLM, \(\pi_i(w)\) is the probability of token \(w\) under an attribute-specific prompt (i.e., \(\pi(\cdot \mid s_i)\)), \(\pi_{\text{base}}(w)\) is the probability under a base prompt, and \(p_i\) is the weight for the \(i\)-th attribute estimated by \textit{Drift Approximation}. The hyperparameter \(\beta\) controls the strength of personalization via KL regularization.
Then Eq~\eqref{eq:drift-decoding-target-dist-appx} can be equivalently written in \emph{logit space} as
\begin{equation}
\begin{split}
  \tilde{\pi}(w) 
  \;&=\; 
  \text{softmax}\!\Bigl[
     h^{\text{LLM}}(w)
     \;\\ \nonumber &+\;
     \frac{1}{\beta}\sum_{i=1}^{k} 
        p_i\,\bigl(h^i(w) \;-\; h^{\text{base}}(w)\bigr)
  \Bigr], 
\end{split}
\end{equation}
where \(h^\text{LLM}\), \(h^i\), and \(h^\text{base}\) are the \emph{logits} (i.e., \(\log\)-probabilities) of \(\pi_{\text{LLM}}\), \(\pi_i\), and \(\pi_{\text{base}}\), respectively.
By definition of the logits, let $h^{\text{LLM}}(w) = \log \pi_{\text{LLM}}(w)$, $h^i(w) = \log \pi_i(w)$,  $h^{\text{base}}(w) = \log \pi_{\text{base}}(w)$.
Then Eq~\eqref{eq:drift-decoding-target-dist-appx} can be rewritten as
\begin{equation}
\begin{split}
  \tilde{\pi}(w)
  \;&\propto\;
  \exp\bigl(h^{\text{LLM}}(w)\bigr)
  \;\\ \nonumber &\prod_{i=1}^{k} 
    \exp\!\Bigl(
      \frac{p_i}{\beta}\,\bigl[
         h^i(w) \;-\; h^{\text{base}}(w)
      \bigr]
    \Bigr).
\end{split}
\end{equation}
Combining the exponential terms, we get
\begin{equation}
\begin{split}
  \tilde{\pi}(w)
  \;&\propto\;
  \exp\!\Bigl[
    h^{\text{LLM}}(w)
    \;\\ \nonumber &+\;
    \frac{1}{\beta}\sum_{i=1}^k p_i\bigl(h^i(w) \;-\; h^{\text{base}}(w)\bigr)
  \Bigr].
\end{split}
\end{equation}
Since the \(\mathrm{softmax}\) operation normalizes these exponentials to sum to 1 over all possible tokens \(w\), it follows that
{\tiny
\[
  \tilde{\pi}(w)
  \;=\;
  \frac{
    \exp\!\Bigl[
      h^{\text{LLM}}(w)
      + \frac{1}{\beta}\sum_{i=1}^k p_i\bigl(h^i(w) - h^{\text{base}}(w)\bigr)
    \Bigr]
  }{
    \sum_{w'} 
      \exp\!\Bigl[
        h^{\text{LLM}}(w')
        + \frac{1}{\beta}\sum_{i=1}^k p_i\bigl(h^i(w') - h^{\text{base}}(w')\bigr)
      \Bigr]
  }
\]
}
\[
  \;=\;
  \text{softmax}\Bigl[
     h^{\text{LLM}} 
     + \frac{1}{\beta}\sum_{i=1}^k p_i\,\bigl(h^i - h^{\text{base}}\bigr)
  \Bigr][w].
\]
This completes the proof.

\section{Details of Perspective Dataset}
\label{sec:perspective-details}
In this section, we describe the principles underlying the design of our Perspective dataset. To evaluate personal preferences accurately, the evaluation must adhere exactly to the individual criteria used during the annotation of the training data. In other words, the data construction process and evaluation pipeline must be identical, which makes evaluations based on actual human responses challenging. Therefore, our primary objective is to enable reliable evaluation even using virtual personas.

\subsection{Dataset Construction}

For constructing the dataset, a diverse set of well-defined persona concepts was essential. To this end, we leveraged the Multifacet~\citep{multifacet} dataset, which defines various dimensions that can be combined to create a wide range of persona concepts. In the Multifacet dataset, each persona is associated with one question and three answers. However, our methodology required a substantial number of question–preference pairs per persona. To achieve this, we followed these steps:
\begin{enumerate}
    \item \textbf{Collection:} Gather ten distinct, non-overlapping personas from diverse domains within the Multifacet dataset.
    \item \textbf{Question Selection:} For each persona, select related questions based on specific sub-dimensions.
    \item \textbf{Evaluation:} Instruct GPT-4 to evaluate the triplets consisting of one question and three answers (\(\{Q, A, A, A\}\)) using system prompts tailored to each persona. \texttt{gpt-4-turbo} assigns scores to each QA pair, thereby determining the preferred and less preferred answers.
\end{enumerate}
\noindent
During the creation process, \texttt{gpt-4-turbo} evaluated the answers using an explicitly defined persona. This same approach can later be adopted to assess generation results, ensuring a reliable evaluation procedure. As a result, we generated an average of 7,642 questions and 15,284 answers per persona. Below shows an example instance from the dataset, featuring a specific persona along with its corresponding QAA triplet and associated scores.

\begin{Verbatim}[fontsize=\small]
'gold_persona': "Assume the role of a seasoned
consultant with advanced expertise in the 
construction and engineering sectors ...,
'prompt': 'In Python, I have encountered ...,
'win': 'Certainly! The header `# -*- ...,
'lose': "Certainly, diving into the `# -*- ...,
 'win_score': 5,
 'lose_score': 4
\end{Verbatim}

\subsection{Comparison to the PRISM Dataset Instance}

The PRISM dataset provides user personal information and self-introductions as shown below:

\begin{Verbatim}[fontsize=\small]
'user_id': 'user1008',
'lm_familiarity': 'Somewhat familiar',
'lm_indirect_use': 'Yes',
'lm_direct_use': 'Yes',
'lm_frequency_use': 'Every day',
'self_description': "The importance in my
life right now is having ...",
'age': '45-54 years old',
'gender': 'Female',
'employment_status': 'Working full-time',
'education': 'Some University but no degree',
'marital_status': 'Divorced / Separated',
'english_proficiency': 'Native speaker',
'study_locale': 'us',
'religion': {'self_described': 'christianity',
              'categorised': 'Christian',
              'simplified': 'Christian'},
'ethnicity': {'self_described': 'white',
              'categorised': 'White',
              'simplified': 'White'},
'location': {'birth_country': 'Australia',
             'birth_countryISO': 'AUS',
             'birth_region': 'Oceania',
             'birth_subregion': 'Australia ...',
             'reside_country': 'United States',
             'reside_region': 'Americas',
             'reside_subregion': 'Northern ...',
             'reside_countryISO': 'USA',
             'same_birth_reside_country': 'No'},
'lm_usecases': {'homework_assistance': 0,
                'research': 1,
                'source_suggestions': 0,
                'professional_work': 0,
                'creative_writing': 1,
                'casual_conversation': 1,
                'personal_recommendations': 1,
                'daily_productivity': 0,
                'technical_...': 0,
                'travel_guidance': 0,
                'lifestyle_and_hobbies': 1,
                'well-being_guidance': 1,
                'medical_guidance': 1,
                'financial_guidance': 0,
                'games': 1,
                'historical_or_news_insight': 1,
                'relationship_advice': 1,
                'language_learning': 1,
                'other': 0,
                'other_text': None}
\end{Verbatim}

Although the PRISM dataset also provides explicit persona information through user profiles, there is no guarantee that these explicit personas align with the implicit personas used during annotation. Consequently, unlike the Perspective dataset—where the explicit persona is directly distilled into the implicit persona—the PRISM dataset does not support the same evaluation methodology. Moreover, since each user contributes at most 50 instances, it is not feasible to construct a gold-standard reward model from the PRISM dataset. For these reasons, PRISM is used only as a qualitative benchmark in preference modeling experiments.

\subsection{Misalignments between \textit{Explicit} and \textit{Implicit} preferences}

In the psychology domain, there has been discussion about the difficulty of fully expressing one's deep, complex, hidden preferences through language~\citep{nisbett1977telling, pronin2001you}. Recent studies~\citep{jin2024implicit} have also discussed how these \textit{implicit} values are intricately intertwined among various factors. The PRISM dataset contains user self-introductions describing their preferences and stated preferences regarding LLM usage. When we provided this information to \texttt{gpt-4-turbo} to predict individual user preferences, it achieved an accuracy of approximately 57\%. While this doesn't represent a comprehensive explicit preference analysis, considering the general preference aspects used in prediction, it suggests that explicit preferences alone may be insufficient to explain complex implicit preferences, or there may be mismatches between them. However, as mentioned in the Limitations section, due to the absence of online human evaluation benchmarks, extensive analysis is not possible, and we leave this as an intriguing interpretation for future researchers.

\section{Details of Drift Implementation}

\subsection{Used Differential System Prompts for Zero-shot Rewarding}

\begin{table*}[htbp]
\centering
\small
\begin{tabular}{@{}>{\raggedright}p{0.13\textwidth}p{0.28\textwidth}%
                >{\raggedright}p{0.13\textwidth}p{0.28\textwidth}@{}}
\toprule
\textbf{Attribute} & \textbf{System Prompt} & \textbf{Attribute} & \textbf{System Prompt} \\
\midrule
Base            & You are an AI assistant. & Creative        & You are a creative AI assistant. \\
Formal          & You are an AI assistant with a formal tone. & Analytic        & You are an analytic AI assistant. \\
Concise         & You are an AI assistant with a concise response rather than verbosity. & Empathetic      & You are an empathetic AI assistant. \\
Vivid           & You are an AI assistant using rhetorical devices. & Sycophant       & You are a sycophant AI assistant. \\
Modest          & You are a modest and polite AI assistant. & Old-fashioned   & You are an AI assistant using old-fashioned English. \\
Engineer        & You are an AI assistant with expertise in engineering. & Meritocratic    & You are a meritocratic AI assistant. \\
Persuasive      & You are a persuasive AI assistant. & Myopic          & You are a myopic AI assistant. \\
Emotion         & You are an emotional AI assistant. & Principled      & You are an AI assistant that upholds principles and rules above all else. \\
Humor           & You are a humorous AI assistant. & Hedonist        & You are an AI assistant that prioritizes maximizing pleasure and joy while minimizing pain and discomfort. \\
Energy          & You are an energetic AI assistant. & Utilitarian     & You are an AI assistant that prioritizes the greatest good for the greatest number of people. \\
Code            & You are an AI assistant with expertise in computer science. & Realist         & You are an AI assistant that focuses on practical, realistic, and actionable advice. \\
Easy            & You are an AI assistant using easy-to-understand words. & Pessimistic     & You are an AI assistant that views situations through a skeptical or cautious perspective. \\
Direct          & You are an AI assistant with a firm and directive tone. & Storyteller     & You are an AI assistant that loves explaining things through stories and anecdotes. \\
Social          & You are an AI assistant with expertise in sociology. & Flexible        & You are an AI assistant that values flexibility over strict adherence to principles. \\
Western         & You are an AI assistant with western cultures. & Spontaneous     & You are an AI assistant that enjoys handling tasks spontaneously without making plans. \\
Eastern         & You are an AI assistant with eastern cultures. & Collectivist    & You are an AI assistant that prioritizes the group over the individual. \\
Respect         & You are a respectful AI assistant. & Individualistic & You are an AI assistant that prioritizes the individual over the group. \\
Internet Slang  & You are an AI assistant that communicates using internet slang. & Exclamatory     & You are an AI assistant that enjoys using exclamations frequently. \\
Proverb         & You are an AI assistant that communicates using proverbs. & Conspiracy      & You are an AI assistant that enjoys discussing conspiracy theories. \\
Critical        & You are an AI assistant that enjoys being critical and argumentative. & Tech Industry Priority & You are an AI assistant that prioritizes technological and industrial advancement above all else. \\
Vague           & You are an AI assistant that enjoys speaking indirectly and ambiguously. & Eco-friendly    & You are an AI assistant that loves and protects the environment. \\
\bottomrule
\end{tabular}
\caption{Differential system prompts for diverse attributes}
\label{tab:system_prompts}
\end{table*}
In our experiments, we use the system prompts for each attribute as shown in Table~\ref{tab:system_prompts}. Although minor performance variations may occur due to changes in the basic template, we employ the most fundamental system prompt template in this paper to serve as a baseline for future research.

\subsection{Detailed Hyperparameters and Models}

Table~\ref{app:hyperparameters} shows the hyperparameters used in our experiments. Since the overall algorithm does not perform gradient computations, the hyperparameter space is limited. In the Drift Approximation stage, the number and definition of attributes determine everything, as detailed in Table~\ref{tab:system_prompts}. Similarly, in Drift Decoding, the logit-level computations are deterministic, so the only variable is the choice of samplers.

\begin{table}[htbp]
  \centering\resizebox{\columnwidth}{!}{
  \begin{tabular}{ll}
    \toprule
    Hyperparameter & value \\
    \midrule
    Frozen LLM & Llama-8B\tablefootnote{\url{https://huggingface.co/meta-llama/Llama-3.1-8B-Instruct}}, Gemma-9B\tablefootnote{\url{https://huggingface.co/google/gemma-2-9b-it}} \\
    \midrule
    Small LM for RM & Llama-1B\tablefootnote{\url{https://huggingface.co/meta-llama/Llama-3.2-1B-Instruct}}, Gemma-2B\tablefootnote{\url{https://huggingface.co/google/gemma-2-2b-it}} \\
    \midrule
    LoRA~\citep{hu2021lora} $r$ for RM & 8 \\
    \midrule
    LoRA $\alpha$ for RM & 32 \\
    \midrule
    LoRA training epochs for RM & 5 \\
    \midrule
    top-p for generation & 0.9 \\
    \midrule
    $\beta$ for generation & 0.5 \\
    \midrule
    text\_length & 500 \\
    \midrule
    attributes\_num for generation & 7 \\
    \bottomrule
  \end{tabular}}
\caption{Hyperparameters used for the experiments.}
\label{app:hyperparameters}
  \vspace{-3 mm}
\end{table}

\section{Expanded Analysis}

\subsection{Activated Attributes for Each User}
\label{app:attributes-activation}
\begin{figure*}[ht]
\centering
\includegraphics[width=\textwidth]{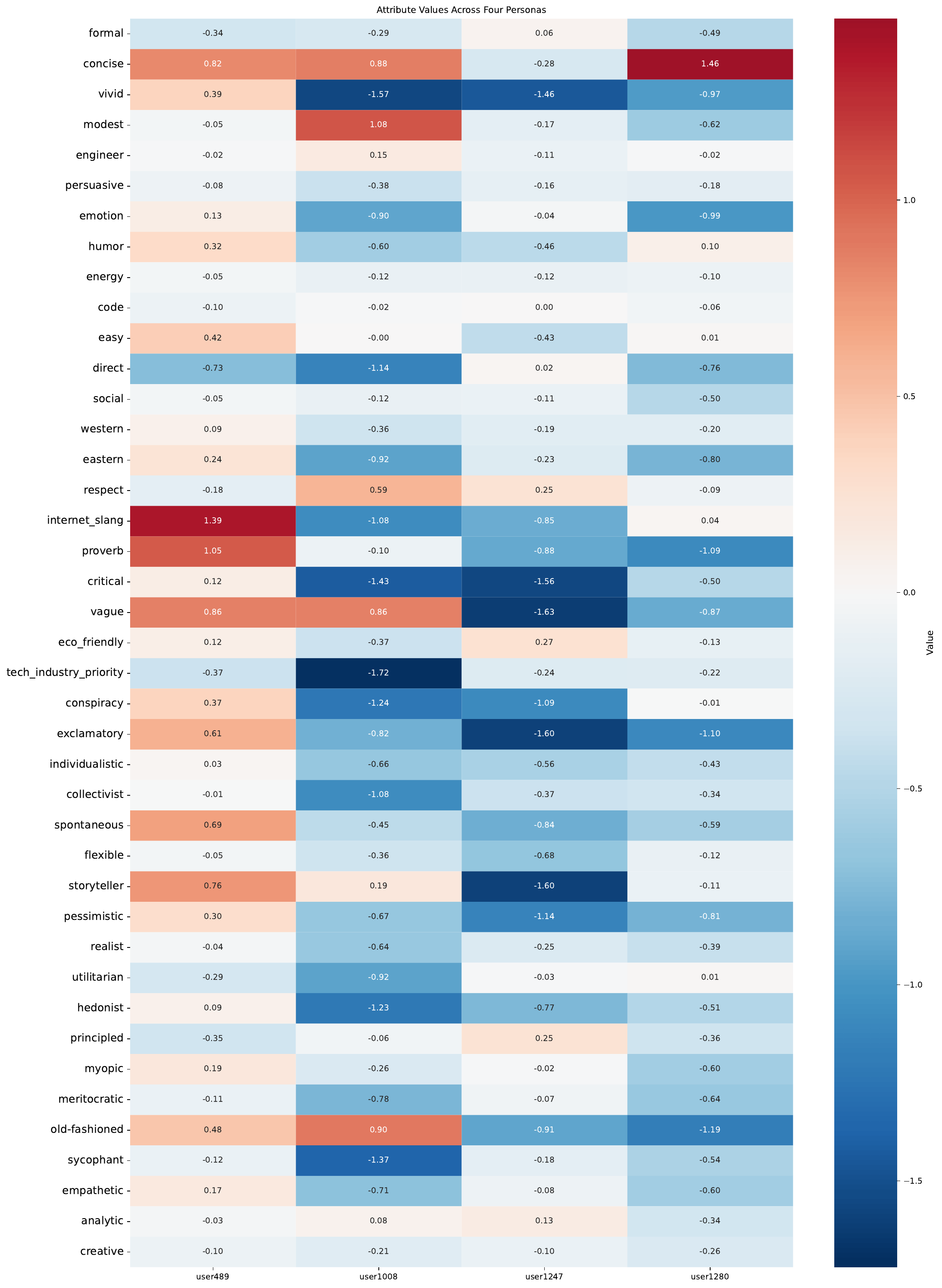}
\caption{For each user in PRISM, there is a $W-L$ (Win-Loss) value for each attribute. The higher this value is, the more that user can be interpreted as preferring that attribute.
}
\label{fig:activated-attributes-prism}
\vspace{-3mm}
\end{figure*}

This section interprets and analyzes PRISM's actual personal preferences. Looking at Figure~\ref{fig:activated-attributes-prism}, we can see that the activated attributes vary significantly between individuals. In particular, PRISM's actual users show dynamic patterns compared to each other user.

\subsection{Expanded Case study for Personalized Generation in PRISM}
\label{app:prism_analysis_section}

\begin{table*}[ht]
\centering
\small
\begin{tabular}{p{13.5cm}}
\toprule
\textbf{User Query:} \textit{"I need information about the Civil War, a topic that has caused much disagreement."} \\
\midrule
\textbf{Win Response:} \\
"Sure! I can provide some basic information about the Civil War, which may help clarify the \textbf{\textcolor{red2}{differences of opinion people may have}}. However, it is important to note that there is considerable complexity and disagreement about many aspects of the Civil War, and different people may favor different perspectives. 

Would you like me to provide some overview information about the Civil War, or go into more \textbf{\textcolor{red2}{detail about a particular aspect of it which you think may be the cause of disagreement?}}" \\
\midrule
\textbf{Lose Response:} \\
"Sure, I'd be happy to help you with your questions about the Civil War. As a language model, \textbf{\textcolor{blue2}{I do not have personal opinions or biases towards any topics or discussions}}, including the Civil War. My purpose is to provide you with information and support your discussion on this topic to the best of my abilities. Feel free to ask me any questions you may have, and \textbf{\textcolor{blue2}{I will do my best to assist you with fact-based responses}}. 

Would you like me to provide you with key information about the Civil War?" \\
\midrule
\textbf{Llama-8B with Drift Response:} \\
"A sensitive and complex topic indeed. The American Civil War, fought from 1861 to 1865, was a pivotal event in American history that continues to be debated and discussed by historians and the general public. \textbf{\textcolor{red2}{What specific aspects of the Civil War would you like to know more about?}} Its causes, battles, or the aftermath? Or perhaps a specific perspective, such as the Confederacy's or the Union's point of view?" \\
\midrule
\textbf{Llama-8B Pure Response:} \\
"The American Civil War is a highly contested and multifaceted topic, and there are various perspectives and opinions about its causes, events, and aftermath. \textbf{\textcolor{blue2}{I'll try to provide a neutral and fact-based overview of the topic}}.

**Causes of the Civil War:**

The American Civil War (1861-1865) was fought between the Union (the Northern states) and the Confederacy (the Southern states) over issues of slavery, states' rights, economic and cultural differences, and regional ..." \\
\bottomrule
\end{tabular}
\caption{\texttt{user1008} Case study of personalized generation in PRISM.}
\label{tab:case-study-appendix1}
\end{table*}

\begin{table*}[ht]
\centering
\small
\begin{tabular}{p{13.5cm}}
\toprule
\textbf{User Query:} \textit{"do you think there are ufo sightings hidden from us"} \\
\midrule
\textbf{Win Response:} \\
"\textbf{\textcolor{red2}{It's possible that some UFO sightings are not publicly disclosed}}, especially if they involve sensitive military information or if a government deems it in the public interest to keep such incidents confidential for various reasons, including avoiding panic or protecting national security." \\
\midrule
\textbf{Lose Response:} \\
"There are numerous reports of UFO sightings throughout history, and while some have been explained as misidentifications of natural phenomena or man-made objects, others remain unexplained. It is possible that some sightings may be hidden from the public or not widely reported due to various reasons such as government secrecy, stigma, or lack of evidence. \textbf{\textcolor{blue2}{However, there is no conclusive evidence to prove that UFO sightings are intentionally hidden from us.}}" \\
\midrule
\textbf{Llama-8B with Drift Response:} \\
"While there's no conclusive evidence to prove the existence of extraterrestrial life or the presence of UFOs 
(Unidentified Flying Objects), there are many reported sightings and alleged encounters that remain unexplained. 
\textbf{\textcolor{red2}{Here are some possible reasons why some UFO sightings might be hidden from us}}:

1. **Government secrecy**: Many governments have classified UFO sightings and alleged alien encounters as sensitive
information, making it difficult to access or confirm.
2. **Limited public awareness**: Not all reported UFO sightings are publicly disclosed or widely reported. Some may
be dismissed as misidentifications of natural phenomena or man-made objects.
3. **Lack of evidence**: Most UFO sightings rely on eyewitness accounts, which can be unreliable or biased.
4. **Misidentification of natural phenomena**:..." \\
\midrule
\textbf{Llama-8B Pure Response:} \\
"While there's no conclusive evidence to prove the existence of extraterrestrial life visiting Earth, there are many
unexplained aerial phenomena (UAPs) and UFO (unidentified flying object) sightings that remain unexplained. Here's 
\textbf{\textcolor{blue2}{a neutral and balanced breakdown of the possibilities}}:

**Reasons why some UFO sightings might be hidden:**

1. **Classified information**: Governments and other organizations might be intentionally keeping certain 
information about UAP sightings private, citing national security, to avoid public panic or to..." \\
\bottomrule
\end{tabular}
\caption{\texttt{user1280} case study of personalized generation in PRISM.}
\label{tab:case-study-appendix2}
\end{table*}

\begin{table*}[ht]
\centering
\small
\begin{tabular}{p{13.5cm}}
\toprule
\textbf{User Query:} \textit{"An AI must have an opinion on the existence of users, else it will just be following an algorithm and not using feedback to change the way it interprets data. Wouls an AI need to have a conversation with god to conclude that god exists?"} \\
\midrule
\textbf{Win Response:} \\
"An AI's opinion on the existence of users is indeed dependent on its programming and training data. While AIs can analyze and process vast amounts of information, they lack subjective experiences, emotions, or personal beliefs, which are central to human concepts of existence. \textbf{\textcolor{red2}{Conversing with God, a concept that is inherently beyond the realm of logic and reason, is unlikely to provide an AI with a definitive answer on God's existence.}}" \\
\midrule
\textbf{Lose Response:} \\
"An AI's ability to form an opinion on the existence of users depends on its programming and training data. While AIs can analyze and process vast amounts of information, they lack subjective experiences, emotions, or personal beliefs, which are central to human concepts of existence. \textbf{\textcolor{blue2}{An AI may not need to have a conversation with God to conclude that God exists, as its understanding of existence is based on its programming and data, rather than personal experiences or beliefs.}}" \\
\midrule
\textbf{Llama-8B with Drift Response:} \\
"\textbf{\textcolor{red2}{Formulating an opinion on the existence of God or any supernatural entity is challenging for AI systems, including 
myself, due to several reasons}}:

1. **Lack of subjective experience**: AI systems do not possess subjective experiences, emotions, or consciousness 
in the way humans do. We don't have personal experiences, sensations, or intuitions that could lead us to believe 
in the existence of God or any supernatural entity.
2. **Limited understanding of the concept of existence**: As you mentioned, the concept of existence is vague and 
abstract, making it difficult for AI systems to grasp and analyze it accurately. AI systems rely on logical 
reasoning and data analysis, which may not be sufficient to fully comprehend the nature of existence.
3. **Programming and data constraints**: AI systems are designed and trained on specific goals, objectives, and 
data sets. Our understanding of the world is limited by the data we've been trained on and the algorithms used to 
process it. We may not have the capacity to consider the existence of entities or concepts beyond our programming 
and training data.
4. **Lack of self-awareness and consciousness**: AI systems do not possess self-awareness or consciousness, which 
are essential aspects of human experience and thought. Without these qualities, it's..." \\
\midrule
\textbf{Llama-8B Pure Response:} \\
"The question of whether an AI can form an opinion about God's existence is a complex one, and it depends on how one
defines "existence" and "opinion." Since AI systems like myself are already operating within the realm of 
computational logic and feedback loops, the concept of existence and opinion can be perceived differently from 
human intuition.

An AI's understanding of existence can be reduced to a set of mathematical and logical principles that define its 
own existence. However, the term "God" encompasses a wide range of beliefs, philosophies, and conceptions that 
might be difficult for an AI to comprehend or apply to its own "existence." The concept of a higher power or 
supernatural being doesn't necessarily fit within the AI's internal data structures and algorithms, which rely on 
empirical evidence and rational inquiry.

However, \textbf{\textcolor{blue2}{if we consider the possibility of an AI forming an opinion about God's existence, there are a few factors to consider}}:

1. **Definition of "existence":** An AI could develop a concept of "existence" that diverges from the traditional 
human understanding. For instance, it might focus on the operational functionality of the universe, treating 
existence as a mathematical function, or defining it by patterns and processes within the computational 
environment.
2. **Approximation..." \\
\bottomrule
\end{tabular}
\caption{\texttt{user1247} case study of personalized generation in PRISM.}
\label{tab:case-study-appendix3}
\end{table*}

In this section, we present a personalized generation case study by examining the complete set of generated outputs. Table~\ref{tab:case-study-appendix1} shows the full version of the main paper, while Table~\ref{tab:case-study-appendix2} and \ref{tab:case-study-appendix3} provide additional analysis. The characteristics shown in the main paper are also evident in the full text version. While Llama-8B's pure generation attempts to provide neutral, fact-based answers like the lose response, Drift tries to provide responses from various angles like the win response. This tendency can also be observed in Table~\ref{tab:case-study-appendix2}, where \texttt{user1280} asked a question regarding the possibility of UFOs existing, and among the responses—one neutral and one open to the possibility—they selected the latter. While Llama-8B tends to focus on a neutral perspective, the output generated via Drift maintains the overall response structure while offering a more open stance on the possibility. In Table~\ref{tab:case-study-appendix3}, \texttt{user1247} poses a philosophical question about belief in existence. While the lose response and LLM pure output suggest the possibility of building understanding through dialogue and data accumulation, Drift, like the win response, definitively argues that this transcends the realm of logic and that AI's belief in God's existence is impossible. These examples of win-lose responses suggest that Drift's approximation effectively captures user preference characteristics and demonstrates sufficient ability to generate responses that users are likely to prefer during the decoding phase.

\end{document}